\documentclass[letterpaper, 10 pt, conference]{ieeeconf}  

\IEEEoverridecommandlockouts                              

\usepackage{cite}
\usepackage{amsmath,amssymb,amsfonts}
\usepackage{algorithmic}
\usepackage{graphicx}
\usepackage{textcomp}
\usepackage[dvipsnames,x11names]{xcolor}

\usepackage{times}
\usepackage{epsfig}
\usepackage{graphicx}
\usepackage{amsmath}
\usepackage{amssymb}
\usepackage{textcomp}
\usepackage{multirow}
\usepackage{hhline}
\usepackage{subcaption}
\usepackage{verbatim}
\usepackage{tabularx}
\usepackage[group-separator={,},mode=text]{siunitx}

\let\labelindent\relax
\usepackage{enumitem}

\usepackage{gensymb}
\usepackage{flushend}
\usepackage{xcolor}

\usepackage[ruled,vlined]{algorithm2e}

\usepackage[labelformat=simple]{subcaption}
\DeclareCaptionLabelSeparator{periodspace}{.\quad}
\definecolor{green1}{RGB}{14, 81, 7}
\definecolor{green2}{RGB}{63, 125, 49}
\definecolor{green3}{RGB}{123, 175, 112}

\renewcommand{\thetable}{\arabic{table}}
\usepackage{booktabs}
\usepackage{array}

\def\BibTeX{{\rm B\kern-.05em{\sc i\kern-.025em b}\kern-.08em
    T\kern-.1667em\lower.7ex\hbox{E}\kern-.125emX\}}}

\makeatletter
\let\NAT@parse\undefined
\makeatother
\usepackage{hyperref}
\hypersetup{
  colorlinks=true,
  citecolor=SpringGreen4
}
\usepackage{cleveref}

\title{\LARGE \bf
WoodScape Motion Segmentation for Autonomous Driving \\ -- CVPR 2023 OmniCV Workshop Challenge
}

\author{Saravanabalagi Ramachandran$^{1}$, S. Nathaniel Cibik$^{2}$, Ganesh Sistu$^{3}$, John McDonald$^{1}$
\thanks{$^{1}$Saravanabalagi Ramachandran and John McDonald are with Lero - the Irish Software Research Centre and the Department of Computer Science, Maynooth University, Maynooth, Ireland {\tt \{saravanabalagi.ramachandran, john.mcdonald\}@mu.ie}.}
\thanks{$^{2}$S. Nathaniel Cibik is with Parallel Domain, USA. {\tt \{nate.cibik \}@gmail.com}.}
\thanks{$^{3}$Ganesh Sistu is with Valeo Vision Sytems, Ireland. {\tt \{ganesh.sistu \}@valeo.com}.}
}
\begin{document}
\maketitle
\thispagestyle{empty}
\pagestyle{empty}
\bstctlcite{IEEEexample:BSTcontrol}
\begin{abstract}
Motion segmentation is a complex yet indispensable task in autonomous driving. The challenges introduced by the ego-motion of the cameras, radial distortion in fisheye lenses, and the need for temporal consistency make the task more complicated, rendering traditional and standard Convolutional Neural Network (CNN) approaches less effective. The consequent laborious data labeling, representation of diverse and uncommon scenarios, and extensive data capture requirements underscore the imperative of synthetic data for improving machine learning model performance. To this end, we employ the PD-WoodScape synthetic dataset developed by Parallel Domain, alongside the WoodScape fisheye dataset.
Thus, we present the WoodScape fisheye motion segmentation challenge for autonomous driving, held as part of the CVPR 2023 Workshop on Omnidirectional Computer Vision (OmniCV). 
As one of the first competitions focused on fisheye motion segmentation, we aim to explore and evaluate the potential and impact of utilizing synthetic data in this domain.
In this paper, we provide a detailed analysis on the competition which attracted the participation of 112 global teams and a total of 234 submissions. This study delineates the complexities inherent in the task of motion segmentation, emphasizes the significance of fisheye datasets, articulate the necessity for synthetic datasets and the resultant domain gap they engender, outlining the foundational blueprint for devising successful solutions. Subsequently, we delve into the details of the baseline experiments and winning methods evaluating their qualitative and quantitative results, providing with useful insights.
\end{abstract}

\section{Introduction}

Motion segmentation plays a pivotal role in the realm of autonomous driving, where accurate perception and understanding of dynamic environments are essential for safe and efficient navigation. Autonomous vehicles rely on the identification and segmentation of motion patterns within the visual field to distinguish between moving objects, such as vehicles, pedestrians, and cyclists, and the static elements of the scene. By discerning and/or tracking distinct motion trajectories of various objects within the environment, autonomous driving systems can anticipate and respond to potential hazards, make informed decisions regarding speed, trajectory, and vehicle interactions, ultimately ensuring a level of safety and adaptability that is crucial for the widespread acceptance and success of autonomous driving technology. In this context, we hosted a global challenge for motion segmentation with the aim of exploring the challenges, methodologies, and potential advancements within the domain of autonomous driving, underscoring its significance in enhancing the perceptual capabilities of self-driving vehicles.

The heavy radial distortion of automotive camera sensors adds the difficulties of scale variance, non-linear distortions, and rotations, which are known to pose enormous challenges for standard Convolutional Neural Network (CNN) architectures. Further, in contrast to the relatively straightforward solutions to moving object detection through frame subtraction and differencing techniques, motion segmentation in autonomous vehicles presents a more formidable challenge. The complexity arises from the necessity to account for both the motion of objects within the frame and the ego-motion of the camera itself. In other words, there is no static reference frame to compare with, as the whole visual field is dynamic. Thus, the intricacies of motion segmentation extend beyond more popular computer vision tasks such as object recognition and object detection, incorporating the challenges of temporal consistency and dynamic object behavior. Traditional segmentation methods often struggle with these complexities, failing to capture the nuanced patterns inherent in moving objects across consecutive frames, especially in situations involving ego-motion of the camera. Many existing off-the-shelf models are tailored for static image segmentation, making them sub-optimal for motion-oriented tasks. Therefore, there is a pressing need for advanced computer vision techniques specifically designed to address the unique demands of motion segmentation.

\begin{figure}[tb]
\centering
\includegraphics[width=0.38\columnwidth]{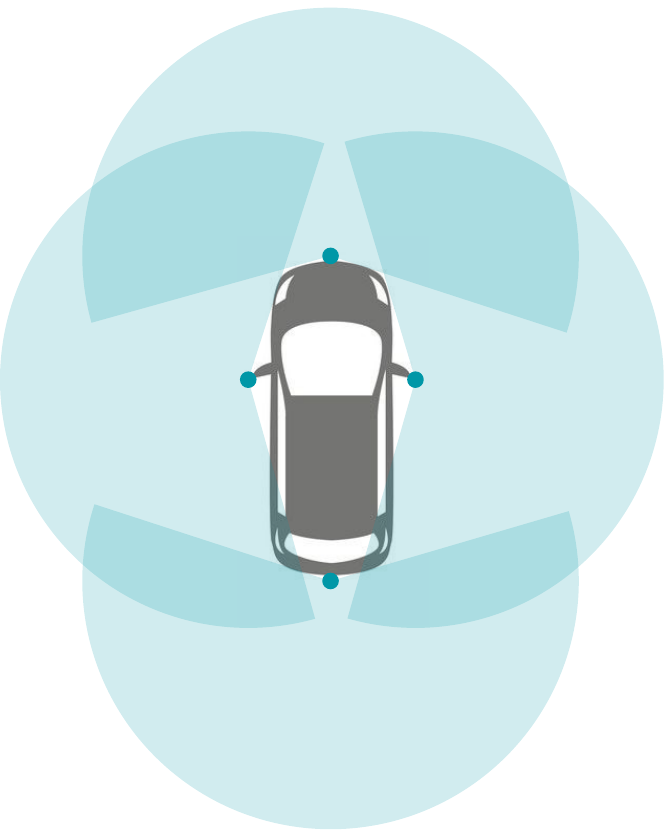}
\includegraphics[width=0.60\columnwidth]{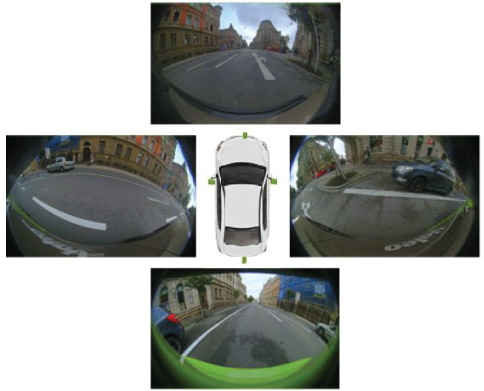}
\caption{Illustration of a typical automotive surround-view system consisting of four fisheye cameras located at the front, rear, and on each wing mirror covering the entire $360^\circ$ around the vehicle. Left: Field of view for each of the four cameras with overlaps. Right: An example frame with images from all the cameras from WoodScape dataset.} 
\label{fig:svs}
\end{figure}

Many tasks in Autonomous Driving can effectively make use of multiple sensors and robust sensor fusion algorithms which combine  sensor information both within and across modalities \cite{yadav2020cnn, mohapatra2021bevdetnet, dasgupta2022spatio}. 
Surround-view systems employ multiple cameras to create a large visual field with large overlapping zones to cover the car's near-field area \cite{eising2021near, kumar2022surround} as shown in \autoref{fig:svs}. For near-field sensing, wide-angle images reaching $180^\degree$ or more are utilized. Any perception algorithm must consider the substantial fisheye distortion that such camera systems produce. Given that most computer vision research relies on narrow field-of-view cameras with modest radial distortion, this is a substantial challenge.

Most commercial cars have fisheye cameras as a primary sensor for automated parking. Rear-view fisheye cameras have become a typical addition in low-cost vehicles for dashboard viewing and reverse parking. Despite their abundance, there are relatively few public databases for fisheye images, which in turn has limited the potential for research in this area. One such dataset is the Oxford RobotCar \cite{maddern20171}, a large-scale dataset with a camera setup consisting of three $180^\degree$ HFoV fisheye cameras facing left, right and rear, and a front-facing perspective trinocular camera with $66^\degree$ HFoV. The principal focus of this dataset, which enables research into continuous learning for autonomous cars and mobile robotics, is localization and mapping. It includes approximately 100 repeated traversals of a continuous route around Oxford, UK, collected over a year and commonly used for long-term localization and mapping. Another dataset that was released more recently is KITTI 360 \cite{kitti360_2022}. It has a camera setup consisting of two $180^\degree$ HFoV fisheye camera facing left and right, and a $90^\degree$ HFoV perspective stereo camera facing the front. Collected in Karlsruhe, Germany, KITTI 360 offers a more comprehensive and diverse dataset, encompassing a wide range of urban and suburban driving conditions and scenarios. The dataset features richer input modalities, comprehensive semantic instance annotations and accurate localization to facilitate research at the intersection of vision, graphics and robotics.

\begin{figure*}[tb]
\centering
\includegraphics[width=\textwidth]{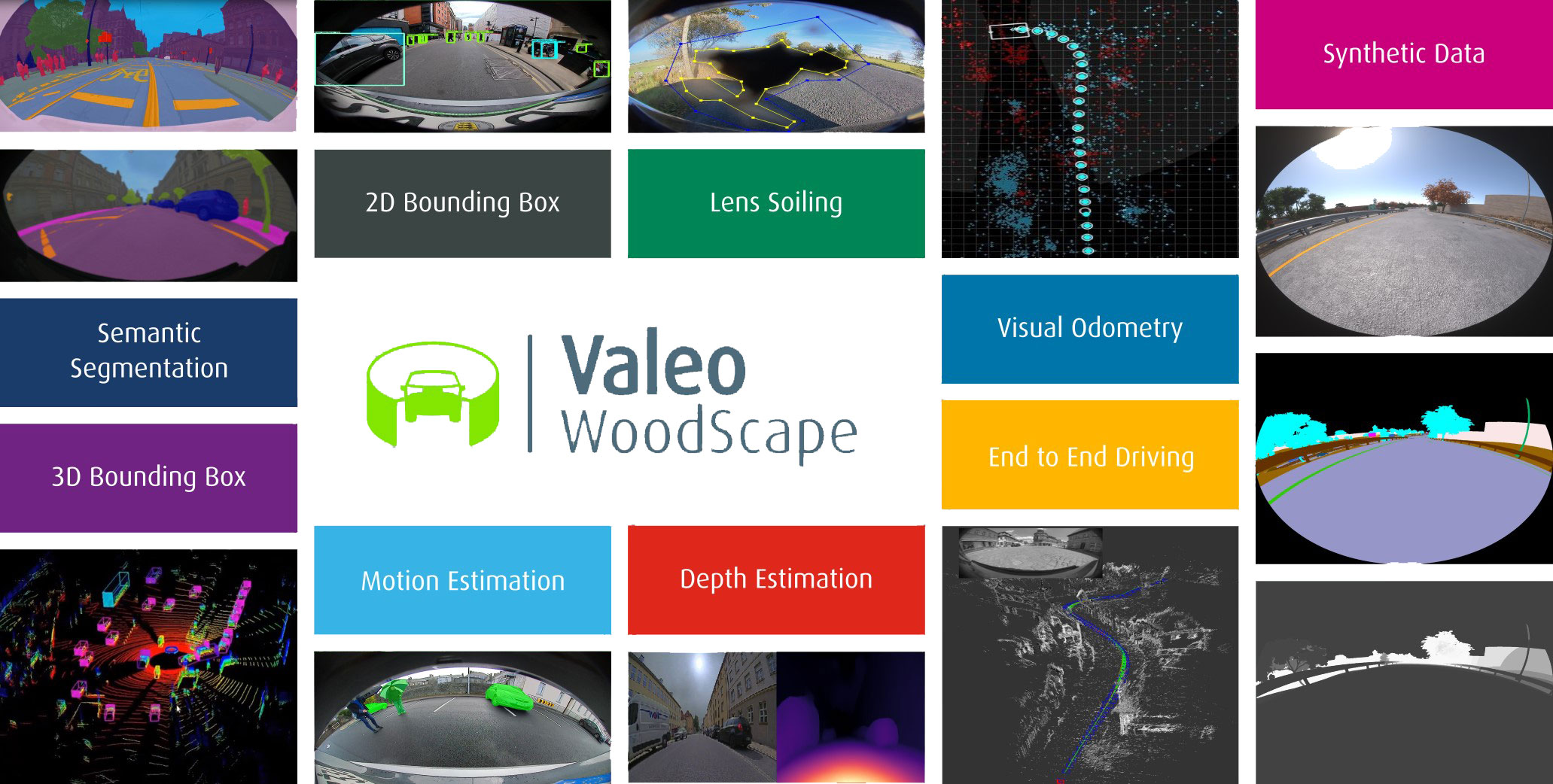}
\caption{Illustration of various perception tasks in WoodScape dataset.} 
\label{fig:woodscape-tasks}
\end{figure*}

WoodScape \cite{yogamani2019_woodscape} is a large dataset for $360^\degree$ sensing around an ego vehicle with four fisheye cameras. It is designed to complement existing automobile datasets with limited FoV images and to encourage further research in multi-task multi-camera computer vision algorithms for self-driving vehicles. It is built based on industrialization needs addressing a diversity of challenges in the field of autonomous vehicles \cite{uricar2019challenges}. The dataset sensor configuration consists of four surround-view fisheye cameras with $190^\degree$ HFoV facing front, rear, left and right as shown in \autoref{fig:svs}. The dataset includes labels for geometry and segmentation tasks, including semantic segmentation, distance estimation, generalized bounding boxes, motion segmentation, and a novel lens soiling detection task (shown in \autoref{fig:woodscape-tasks}). 

Instead of naive rectification, the WoodScape dataset pushes researchers to create solutions that can work directly on raw fisheye images, modeling the underlying distortion. Since it's initial release, the WoodScape dataset has enabled research in various perception areas such as object detection~\cite{dahal2021online, rashedfisheyeyolo, rashed2021generalized,  yahiaoui2019optimization}, trailer detection \cite{dahal2019deeptrailerassist}, soiling detection~\cite{uricar2021let, das2020tiledsoilingnet, uricar2019desoiling}, semantic segmentation \cite{cheke2022fisheyepixpro, sobh2021adversarial, dahal2021roadedgenet, klingner2022detecting, rashed2019motion}, weather classification \cite{dhananjaya2021weather}, depth prediction \cite{kumar2018monocular, kumar2018near, kumar2021svdistnet, kumar2020unrectdepthnet, kumar2020fisheyedistancenet, kumar2021fisheyedistancenet++, kumar2020syndistnet}, moving object detection \cite{siam2018modnet, yahiaoui2019fisheyemodnet, rashed2019motion, mohamed2021monocular}, SLAM \cite{tripathi2020trained, gallagher2021hybrid} and multi-task learning \cite{leang2020dynamic, kumar2021omnidet}.

Despite the scale and extent of the WoodScape dataset, as well as any dataset derived from real-world contexts, ensuring that all features, objects, and agents of interest are represented to desired levels is challenging, and in some cases, impossible. The collection and labeling of large, diverse real-world driving datasets is a tedious and resource-intensive process. Vehicles equipped with sensors must be driven thousands of hours to collect the raw data sequences, which often cover a narrow range of operational domains and fail to provide training examples of edge cases in which the model might fail. The raw data must then be annotated by humans, which is expensive, time consuming, and prone to error.

Synthetic data has long been seen as a potential solution to alleviating the burden of collecting large real-world annotated datasets by generating effective training examples in simulation. The Synscapes dataset \cite{wrenninge2018synscapes} demonstrated that a combination of procedurally generated scenarios, photorealistic rendering techniques, and intentional domain matching to the target Cityscapes dataset \cite{cordts2016cityscapes} led to superior transfer learning compared to previous, less photorealistic synthetic driving datasets. SynWoodScape \cite{sekkat2022synwoodscape} is a synthetic version of the WoodScape dataset. However, like many datasets collected in CARLA, it lacks photorealism, and the study found that joint training on the synthetic and real data did not increase model performance on the target domain without the use of domain adaptation. 

PD-WoodScape was designed by Parallel Domain to match the sensors, annotations, and operational domain of the WoodScape dataset. Rendered with a professional-grade synthetic data pipeline, it demonstrates better photorealism than CARLA, reducing the synthetic-to-real domain gap without domain adaptation. Care was taken to align the annotation spaces between the PD-WoodScape and WoodScape datasets by considering the minimum detection sizes for bounding boxes and masks found in the real dataset, in order to avoid false positives on the real test set that occur when a model has been trained on a perfectly annotated synthetic dataset which includes labels for small or barely visible agents that a human annotator would miss.

Synthetic datasets, while invaluable for training machine learning models, often cannot be directly utilized due to the \textit{domain gap}. This term refers to the disparity between the characteristics of synthetic and real-world data and annotations. Although synthetic datasets are carefully generated through simulations or algorithmic means and are designed to mimic real-world scenarios, they often do not fully replicate the complexity, variability, and occasional unpredictability found in natural data. This gap can lead to models trained on synthetic data performing sub-optimally when applied to real-world tasks, as the learned features and patterns may not align well with those in natural datasets. The domain gap is a critical challenge in the field of machine learning and artificial intelligence, necessitating strategies to bridge or minimize this disparity to enhance the applicability and robustness of models trained on synthetic data. Although the synthetic-to-real domain gap has historically been a hindrance in the transferability of knowledge learned from synthetic data to real-world inference, there is considerable research dedicated to domain adaptation techniques which seek to close this gap \cite{domain_adapt_1_2017, domain_adapt_2_review_10.1007, domain_adapt_3_review_technologies8020035}. 

Advances in rendering capabilities and synthetic data curation techniques can help to minimize the domain gap from the data generation side, making the sensor and annotation spaces more realistic. The domain gap can be ameliorated further after rendering time with post-processing domain adaptation techniques (e.g. image transfer) and training architecture modifications. Color space adaptations \cite{reinhard2001, lyu2020learning} can decrease domain gap while preserving image geometry. GAN-based approaches seek to close the domain gap with adaptations at the pixel level \cite{Imbusch_2022}, the learned feature level \cite{sankaranarayanan2018learning}, or both \cite{hoffman2017cycada}. \cite{ganin2015unsupervised} showed that unsupervised domain adaptation (UDA) in the learned feature space can be achieved through gradient reversal of loss from an added domain classifier head. ADVENT \cite{vu2019advent} achieves UDA via entropy minimization, and FADA \cite{wang2020classes} demonstrated the benefits of class-level feature alignment. The combination of targeted synthetic data generation and domain adaptation techniques has the potential to reduce the need for real training data.

As such, successful solutions to motion segmentation in fisheye images demand an integrated approach that emphasizes training on a combination of real-world and synthetic datasets, complemented by strategies aimed at mitigating the domain gap. 

This paper discusses the third edition of WoodScape dataset challenge focused on the motion segmentation task in autonomous driving leveraging PD-WoodScape dataset. The results of the previous editions of the WoodScape challenge are discussed in \cite{ramachandran2021woodscape, ramachandran2022woodscape}. The challenge was organized as part of the \href{https://sites.google.com/view/omnicv2023}{CVPR 2023 OmniCV workshop}. Section \ref{sec:challenge} discusses the challenge, metrics, and other conditions. Section \ref{sec:outcome} furnishes information regarding participation statistics, baseline model performances, an overview of the top three winning techniques, a tabulated presentation of the scores of top 10 participants, and a brief discussion enriched with critical insights. Finally, Section \ref{sec:conc} provides concluding remarks.

\section{Challenge} \label{sec:challenge}

\begin{figure*}[tb]
\centering
\includegraphics[width=\textwidth]{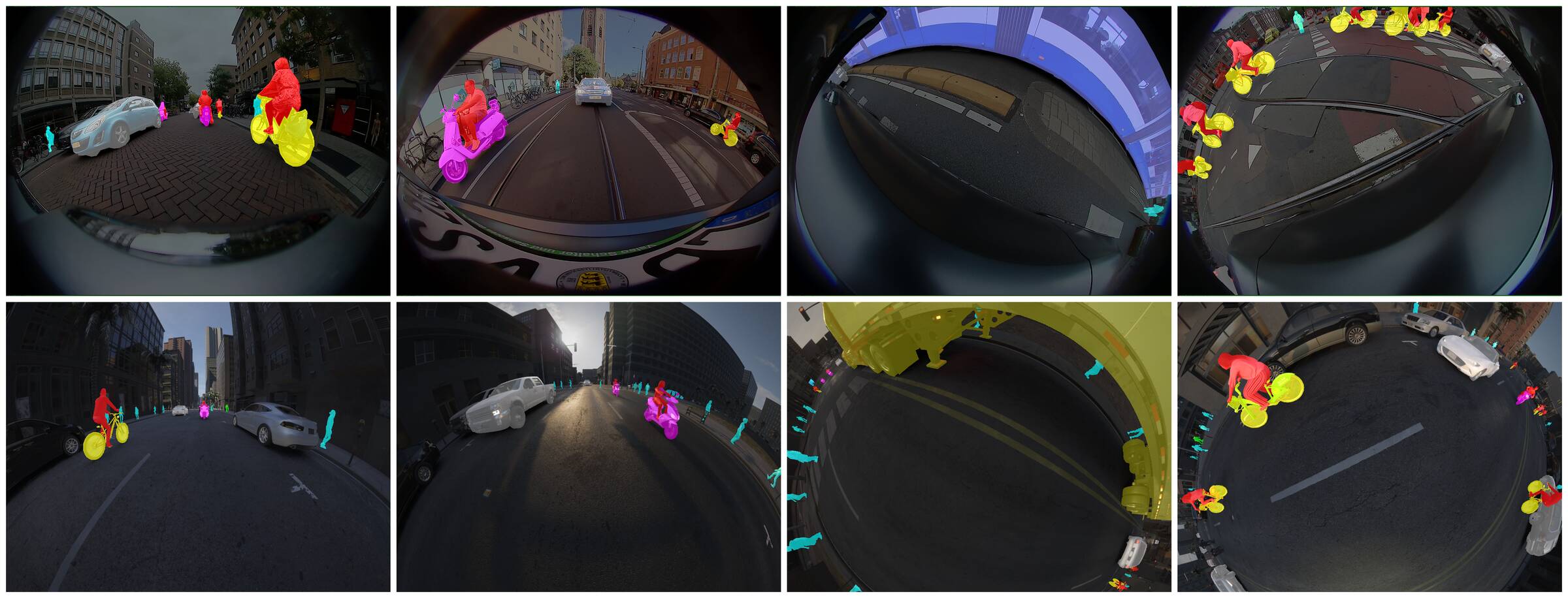}
\caption{Illustration of motion segmentation annotations. Note that while both the real and synthetic dataset offer ground truth with multiple classes, the task for the challenge is a binary classification between static and motion pixels. (Top row: real images from WoodScape dataset. Bottom row: synthetic images from PD-WoodScape dataset)}
\label{fig:segmentation-task}
\end{figure*}

The objective of this challenge is to advance the state of the art in motion segmentation by benchmarking techniques that minimise the quantity of real-world data used during the training process. A visualization of the motion segmentation annotations is seen in \autoref{fig:segmentation-task}. Contestants were given access to both real-world data, through the WoodScape dataset, and synthetic data, through a corresponding  PD-WoodScape dataset that was generated specifically for the challenge. As such, there is a trade-off between the authenticity of real-world data and the scalability and diversity achievable through synthetic data. The challenge, thus, seeks novel techniques that cleverly utilise the complementary nature of these datasets. 

The competition was hosted on CodaLab. The challenge was held in 2 phases: Dev Phase and Test Phase. During the Test Phase, the submissions were evaluated on the entire test set. Details of start and end dates and the duration is shown in table \autoref{tab:challenge_phases}.

The evaluation was done only on the WoodScape dataset. During the Dev Phase, evaluation was carried out on a subset of the original test set to prevent the participants from overfitting their model on the test data. To further impede the participants from modifying their model based on the evaluation scores, the subset was changed every week. However, for any given week, the subset consisting of 1000 images picked at random, was the same for all participants.

\begin{table}
\centering
\begin{tabular}{ccc} \toprule
Split & WoodScape & PD-WoodScape \\\midrule
Training Set & 8234 & 16960 \\
Test Set & 1766 & 0 \\\midrule
Total & 10000 & 16960 \\\bottomrule
\end{tabular}
\caption{Train and test data split of the WoodScape and PD-WoodScape datasets.}
\label{tab:train_test_split}
\end{table}

\begin{table}
\centering
\begin{tabular}{clc} \toprule
Label & Description \\\midrule
0 & Static \\
1 & Animal \\
2 & Grouped Animals \\
3 & Grouped Pedestrian and Animals \\
4 & Rider \\
5 & Person \\
6 & Bicycle \\
7 & Motorcycle \\
8 & Dynamic Car \\
9 & Car \\
10 & Dynamic Van \\
11 & Van \\
12 & Caravan \\
13 & Bus \\
14 & Truck \\
15 & Trailer \\
16 & Train Tram \\
17 & Grouped Vehicles \\
18 & Moveable Objects \\
19 & Other Wheeled Transport \\\bottomrule
\end{tabular}
\caption{Motion Segmentation classes available in the ground truth labels provided.}
\label{tab:label_mapping_multi}
\end{table}

\begin{table}
\centering
\begin{tabular}{ccc} \toprule
Label & Description \\\midrule
0 & Static \\
1 & Motion \\\bottomrule
\end{tabular}
\caption{Binary Motion Segmentation classes the participants are required to submit.}
\label{tab:label_mapping_binary}
\end{table}

\begin{table}
\centering
\begin{tabular}{cccc} \toprule
Phase & Start Date & End Date & Duration \\\midrule
Dev & April 5, 2023 & June 2, 2023 & 59 days \\
Test & June 3, 2023 & June 4, 2023 & 2 days \\\bottomrule
\end{tabular}
\caption{Details of start and end dates of the two phases of the challenge.}
\label{tab:challenge_phases}
\end{table}

\subsection{Datasets}

The WoodScape dataset provides 8234 training images recorded across several European countries through four fisheye cameras, placed in the front, rear, left and right. \autoref{fig:svs} shows the placement of these 4 cameras on the vehicle along with their field of view. With a rich annotation set, it includes urban, highway, and parking scenes in clear weather with variable cloud cover.

The synthetic dataset, PD-WoodScape, comprises 82 urban scenes and 24 parking scenes, captured under variable cloud cover across 4,240 keyframes. Each keyframe provides 2D and 3D bounding boxes, depth, semantic, and motion segmentation annotations recorded through four fisheye cameras, resulting in a total of 16,960 annotated images paired with their previous RGB frames.

The sizes of the real and synthetic dataset are shown in \autoref{tab:train_test_split}. The ground truth labels include a total of 20 classes as listed in \autoref{tab:label_mapping_multi}. While both the synthetic and real datasets have multiclass motion annotations, the challenge focused on a binary classification task of predicting each pixel as either static (class 0) or motion (class 1) as shown in \autoref{tab:label_mapping_binary}. 

\subsection{Metrics}

Intersection over Union (IoU) is a standard evaluation metric for image segmentation tasks, which calculates the ratio between the overlap of predicted and ground truth pixels (the intersection: $\hat{Y} \cap Y$) and their combined area (the union: $\hat{Y} \cup Y$).

\begin{equation}
\text{IoU} = \frac{\text{Area of Intersection}}{\text{Area of Union}}
\end{equation}

Given the significant imbalance between the static and motion classes, we weigh the motion class IoU more heavily to penalise models that over-estimate static class pixels. We set the weights to 0.02 and 0.98 for static and motion pixels respectively, following the measured pixel statistics computed on the WoodScape dataset. 

To encourage techniques which generalize well to the real-world testing set using the least amount of real training data possible, a penalty is applied to this weighted mIoU in proportion to the amount of the real-world dataset used in training. To allow fine-tuning, we allow up to 25\%, corresponding to 2058 images of the real data to be used without any penalty.

\subsection{Reward}
The winning team received €2,500 through sponsorship from Valeo, and the runner up and third place teams received €1,500 and €1,000 respectively from Parallel Domain. The top 3 teams were offered to present in-person or virtually in the OmniCV 4th Workshop held in conjunction with IEEE Computer Vision and Pattern Recognition (CVPR) 2023, held at Vancouver, Canada.

\subsection{Conditions}

Competition rules did not allow the teams to make use of any pre-trained models and other public datasets for pre-training. This was to encourage domain adaptation and fine-tuning techniques utilized by models trained on the given real and synthetic datasets only.
There were no restrictions on computational complexity, training and/or inference time. There was also no limit on team size, however there was a limit of 5 submissions per day and 100 submissions in total perteam.
For the Test phase, teams were limited to a maximum of 2 submissions. 
Valeo employees, Parallel Domain employees, or their collaborators with access to the full WoodScape dataset were not allowed to take part in the challenge. Teams placing in the top 3 were requested to share their code in order for the competition organisers to verify the legitimacy of their approach and their submissions.

\section{Outcome} \label{sec:outcome}

\begin{figure*}[tb]
\centering
\includegraphics[width=\textwidth]{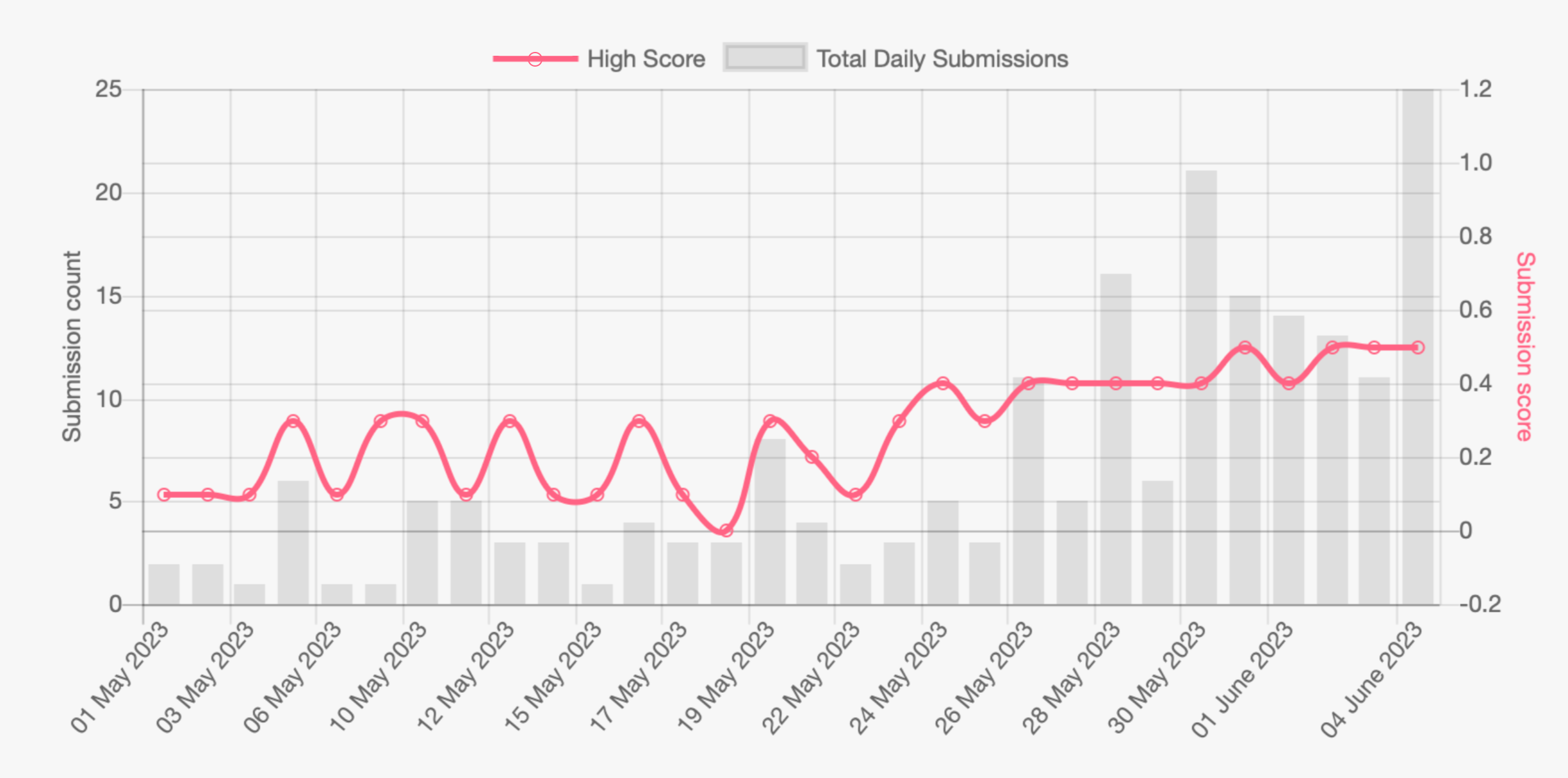}
\caption{Illustration of the trend of number of daily submissions and their scores during the entire phase of the competition.} 
\label{fig:analytics}
\end{figure*}

The competition was active for 61 days from April 5, 2023 through to June 4, 2023. The competition attracted 112 teams worldwide with a total of 234 submissions. An plot of the timeline showing the number of daily submissions and their scores during the entire phase of the competition is shown in \autoref{fig:analytics}.
The submissions and the leaderboard were open from May 1, 2023. It is interesting to note that the final 10 days of the competition attracted over two-thirds of the submissions. This substantial influx of submissions can be attributed to the considerable time required for participants to devise, construct, train, and refine their pipelines in order to generate the desired outcomes, given the intricate nature and complexity of the motion segmentation problem.

\begin{figure*}[tb]
\centering
\includegraphics[width=\textwidth]{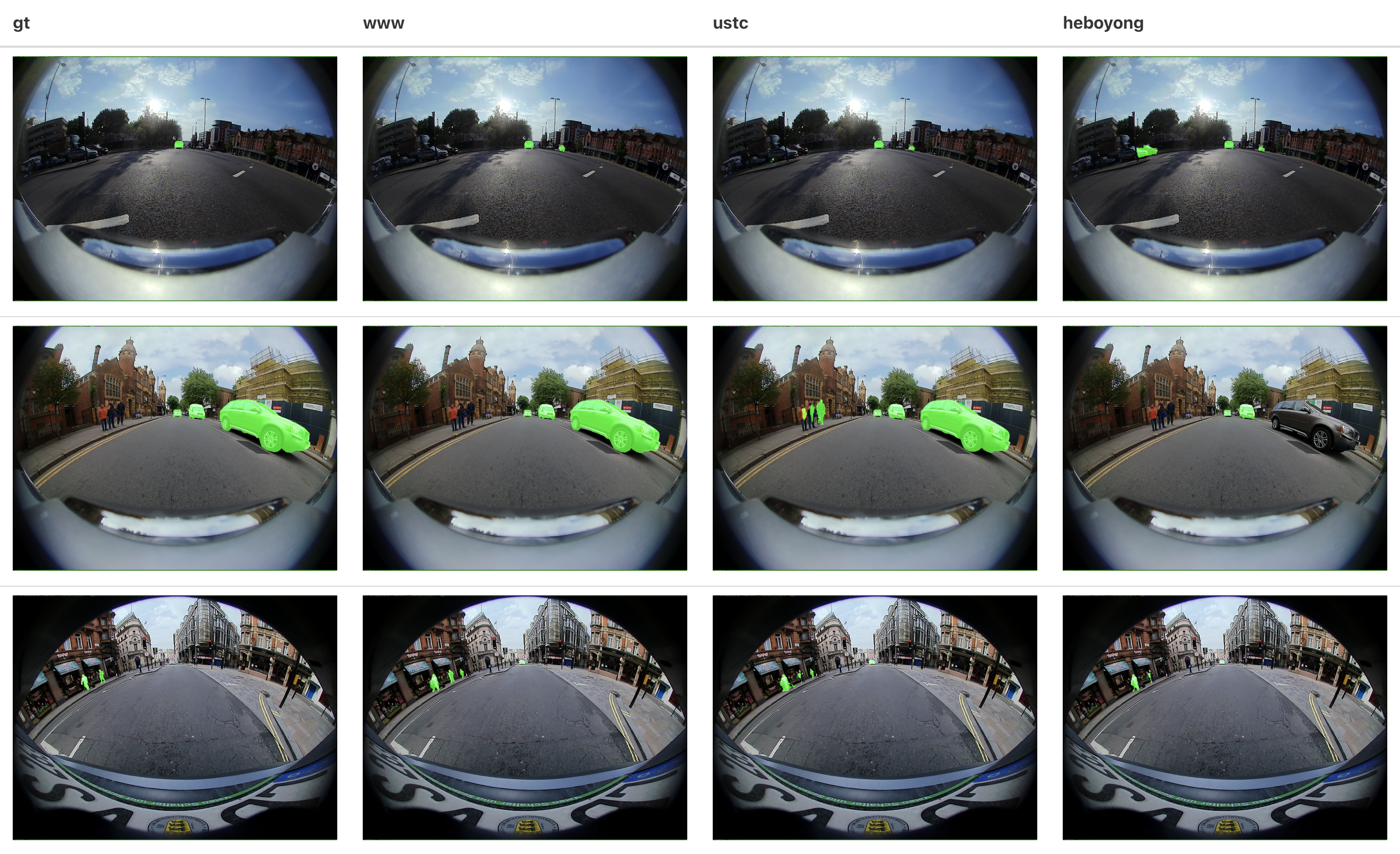}
\caption{Motion segmentation predictions (overlaid on the input image and highlighted in green) by top 3 teams compared with the reference for 3 randomly chosen images. Left to Right: Reference (ground truth labels), Team STAR (winner), Team USTC-IAT-United (second place) and Team XMU-UAV (third place).} 
\label{fig:pred-comparison}
\end{figure*}

\subsection{Baseline}
For the baseline model, a fundamental architecture was chosen utilizing the Resnet-18 \cite{he2015deep_resnet} backbone. The validation was carried out using the WoodScape dataset in line with the evaluation used in the competition. We carry out a set of experiments training multiple instances under controlled settings to enhance our comprehension of how the composition of the training dataset influences the model's performance.

First, we train a baseline model by conducting training on two instances (with different random initial weights) utilizing solely the WoodScape dataset. The resulting IoU Motion scores were calculated to be 0.3780 and 0.3560 (corresponding IoU Static scores being 0.9710 and 0.9700), providing an initial benchmark for the model's predictive accuracy in identifying motion-related features. 
Following this, a subsequent training experiment was conducted, involving three instances trained exclusively on the PD-WoodScape dataset. Intriguingly, the motion scores observed in this setting experienced a notable decrease, settling at values of 0.1500, 0.1720, and 0.1740 for IoU Motion (and 0.9470, 0.9460, and 0.9450 for IoU Static). This decline in performance is due to the unaddressed domain gap, indicating the importance of domain-specific considerations in training. To rectify the impact of this domain gap, three additional instances were trained by leveraging the models previously trained exclusively on the PD-WoodScape dataset. These models were further fine-tuned specifically on the WoodScape dataset. This fine-tuning strategy significantly improved the model's performance on motion classes, resulting in improved IoU Motion scores of 0.3830, 0.4200, and 0.4120 (and IoU Static scores of 0.9700, 0.9720, and 0.9750), higher than the motion scores of models trained exclusively on the WoodScape dataset.

We also experimented WoodScape and PD-WoodScape joint-training where we train 2 instances and obtain motion scores 0.3650 and 0.4010. This only resulted in a marginal increase compared to WoodScape only baselines. This is consistent with the findings of Parallel Domain regarding the domain gap and joint-training experiments \cite{pdtraining2023, cvprsdas2023}. Further, we additionally explored predicting all 20 classes while utilizing loss from all 20 classes, instead of just 2 classes (static and motion). We trained 2 instances and obtain IoU Motion scores of 0.3960 and 0.3270 (and IoU Static scores of 0.9780 and 0.9740).

\subsection{Methods}

\subsubsection{Winning Team}
Team \textit{STAR} (Username \textit{www}) finished in first place with a score of 0.50 (IoU Motion 0.49 and IoU Static 0.99) with their Swin-Large Mask2Former approach. 
The team consisted of Fuxing Leng and Chenlong Yi
affiliated to ByteDance and Huazhong University of Science and Technology, respectively. They used masked attention in the transformer decoder derived from Mask2Former \cite{cheng2021mask2former} coupled with a Swin-Large Swin-Transformer \cite{liu2021swin} backbone. Their technique employed a data augmentation strategy encompassing multi-scale training, random flip, color jitter, and random crop techniques, aimed at enhancing overall generalization performance. They used a two-stage training strategy to achieve superior domain adaptation performance. In the initial phase, a composite training dataset consisting of all PD-WoodScape images (totaling 16,960) and 2,000 WoodScape images is utilized, subjecting the model to 500,000 training iterations. Following this, the model is fine-tuned in the subsequent stage for 200,000 iterations employing a subset consisting only of the first 2,000 WoodScape images. With their technique they achieve the top score of 0.50 without incurring any penalty, resulting in a position of $1^{st}$ place in the competition.

\begin{table*}
\centering
\begin{tabular}{rlllllll}
\toprule
\# & User         & Score  & IoU Static & IoU Motion \\\midrule
{\color{green1} \textbf{1}} & {\color{green1} \textbf{www}}              & {\color{green1} \textbf{0.50 (1)}} & {\color{green1} \textbf{0.99 (2)}} & {\color{green1} \textbf{0.49 (1)}} \\
{\color{green2} \textbf{2}} & {\color{green2} \textbf{USTCxNetEaseFuxi}} & {\color{green2} \textbf{0.47 (2)}} & {\color{green2} \textbf{0.99 (16)}} & {\color{green2} \textbf{0.45 (2)}} \\
{\color{green3} \textbf{3}} & {\color{green3} \textbf{heboyong}}         & {\color{green3} \textbf{0.43 (3)}} & {\color{green3} \textbf{0.99 (3)}} & {\color{green3} \textbf{0.42 (3)}} \\
4                        & bailanteam   & 0.41 (4)    & 0.99 (8)    & 0.40 (4)    \\
5                        & cccc         & 0.41 (5)    & 0.99 (14)   & 0.40 (5)    \\
6                        & bbbb         & 0.40 (6)    & 0.99 (5)    & 0.39 (6)    \\
7                        & ffff         & 0.40 (7)    & 0.99 (9)    & 0.39 (7)    \\
8                        & GenglunZhang & 0.40 (8)    & 0.99 (10)   & 0.39 (8)    \\
9                        & Wizard001    & 0.40 (9)    & 0.99 (15)   & 0.39 (9)    \\
10                       & u7355608     & 0.39 (10)   & 0.99 (6)    & 0.38 (10)   \\\bottomrule
\end{tabular}
\caption{Snapshot of the challenge leaderboard illustrating the top ten participants based on the \textit{Score} metric. Top three participants are highlighted in shades of green.  }
\label{tab:leaderboard}
\end{table*}

\subsubsection{Second Place}
Team \textit{USTC-IAT-United} (Username \textit{USTCxNetEaseFuxi}) finished in second place with a score of 0.49 (IoU Motion 0.45 and IoU Static 0.99) using a mixture of models derived from Mask2Former \cite{cheng2021mask2former}, ConvNeXt \cite{liu_convnext}, InternImage \cite{internimage_Wang_2023_CVPR}, and Segmenter \cite{segmenter_Strudel_2021_ICCV}. The team included 
Jun Yu,
Renjie Lu,
Leilei Wang,
Shuoping Yang,
Gongpeng Zhao,
Renda Li, and
Bingyuan Zhang
each of whom were affiliated to 
the University of Science and Technology of China. 
They employed a model ensemble consisting of the aforementioned four benchmark models and with a pixel based voting strategy to produce the final outputs. In their technique all the input images were uniformly cropped to a constant $640\times640$ and augmentation methods such as multi-scale random resize, random mosaic, random crop, random horizontal flip, and random rotation were employed to improve the robustness of motion segmentation. A cross entropy loss function utilizing all 20 individual motion classes in the ground truth provided was used and the model was trained for 160,000 iterations. The training employed the AdamW \cite{adamw_KingmaB14} optimizer with a PolyLR \cite{polylr_Zhang2020kDecayJA} learning rate, starting with an initial learning rate of 0.0001. The team underscores the significant variability in the performance of the models used, with certain models demonstrating proficiency for specific motion classes, motivating the adoption of a model ensemble approach. This strategic implementation resulted in the team achieving the runner-up position, with a score of 0.47.

\subsubsection{Third Place}
Team \textit{XMU-UAV} (Username \textit{heboyong}) finished in third place with a score of 0.43 (IoU Motion 0.42 and IoU Static 0.99) using UpperNet \cite{upper_xiao2018unified} and K-Net \cite{zhang2021knet} neural networks. The team included
Boyong He,
Weijie Guo,
Xi Lin, and
Yuxiang Ji
affiliated to
Xiamen University. 
Their framework employ ConvNext-Small \cite{liu_convnext} as the backbone. Six distinct models are developed, where three models exclusively employ 25\% WoodScape real data, while the remaining three models incorporate the entirety of PD-WoodScape images along with 25\% WoodScape images. They formulate the final motion segmentation masks as a fusion of outputs of these six models. Data augmentation techniques used for both real data and synthetic data includes multi-scale training, random cropping, rotation augmentation, and photo color augmentation.

The team investigated the exclusive use of synthetic images without any domain adaptation techniques and revealed a performance deficit over the exclusive use of real-world WoodScape images. Furthermore, they found that even mixing synthetic and real images in a similar setting did not match the performance of a real-world only trained model. However, the introduction of domain adaptation methodologies, such as histogram matching, pixel distribution alignment, and Fourier Domain Adaptation \cite{fda_9157228}, serves to mitigate the domain gap between synthetic and authentic datasets, yielding enhanced performance outcomes. They train all the models for 80,000 iterations achieving a score of 0.43 and ranking third in the challenge.

\subsection{Results and Discussion}

Team \textit{STAR}, with a lead score of 0.49 (IoU Motion 0.49 and IoU Static 0.99), was announced as the winner on 12th June 2023. In Table \autoref{tab:leaderboard}, we showcase the challenge leaderboard with details of the top twenty team participants. 
The winning team \textit{STAR} and the runner up team \textit{USTC-IAT-United} presented their method virtually in the OmniCV Workshop, CVPR 2023, held on June 19, 2023 in Vancouver, Canada. In \autoref{fig:pred-comparison}, we illustrate outputs of the top 3 teams from randomly picked sample images. 

The competition revealed several critical insights:

\begin{itemize}
\item \textbf{Attention Mechanisms}: The use of attention mechanisms in transformer architectures hints at their potential utility in segmentation and possibly other computer vision tasks.

\item \textbf{Strategic Training Phases}: Phased training, as demonstrated by top contenders, can be a critical factor in achieving high performance, urging future competitions and research to consider similar approaches.

\item \textbf{Ensemble Approaches}: The success of the ensemble methods from the top teams indicates the potential robustness gained from such approaches, suggesting that future work should focus on ensemble methods as a means to improve model performance and reliability.

\item \textbf{Domain Adaptation}: The third-place team's focus on domain adaptation techniques serves as a reminder that domain-specific challenges should not be overlooked, especially when synthetic and real-world data are involved.
\end{itemize}

\section{Conclusion} \label{sec:conc}
In this paper, we discussed the results of the motion segmentation challenge hosted at our CVPR OmniCV workshop 2023. The competition provided an invaluable platform for researchers and practitioners to tackle the challenging problem of motion segmentation. It highlighted the necessity for domain-specific training, the potential of ensemble methods, and the critical role of domain adaptation techniques. The event also set benchmarks for future endeavors in this domain, revealing the capabilities and limitations of contemporary approaches. The top-performing teams in the competition employed a variety of innovative methods, providing a wealth of insights into effective strategies for tackling the complex problem of motion segmentation. Attention mechanisms, phased training, ensemble methods, and domain adaptation techniques emerged as key factors contributing to high performance. These insights not only set new benchmarks but also provide a roadmap for future research in this rapidly evolving field. We have started accepting submissions again keeping the challenge open to everyone to encourage further research and novel solutions to motion segmentation. In our future work, we plan to organize similar workshop challenges on fisheye multi-task learning.

\section*{Acknowledgments}

WoodScape OmniCV 2023 Challenge
was supported
in part
by \href{https://www.sfi.ie}{Science Foundation Ireland} grant 13/RC/2094 to \href{https://www.lero.ie}{Lero - the Irish Software Research Centre} and grant 16/RI/3399.

\bibliographystyle{IEEEtran}
\bibliography{IEEEabrv, main}

\begin{thebibliography}{10}
\providecommand{\url}[1]{#1}
\csname url@rmstyle\endcsname
\providecommand{\newblock}{\relax}
\providecommand{\bibinfo}[2]{#2}
\providecommand\BIBentrySTDinterwordspacing{\spaceskip=0pt\relax}
\providecommand\BIBentryALTinterwordstretchfactor{4}
\providecommand\BIBentryALTinterwordspacing{\spaceskip=\fontdimen2\font plus
\BIBentryALTinterwordstretchfactor\fontdimen3\font minus
  \fontdimen4\font\relax}
\providecommand\BIBforeignlanguage[2]{{%
\expandafter\ifx\csname l@#1\endcsname\relax
\typeout{** WARNING: IEEEtran.bst: No hyphenation pattern has been}%
\typeout{** loaded for the language `#1'. Using the pattern for}%
\typeout{** the default language instead.}%
\else
\language=\csname l@#1\endcsname
\fi
#2}}

\bibitem{yadav2020cnn}
R.~Yadav, A.~Samir, H.~Rashed, S.~Yogamani, and R.~Dahyot, ``Cnn based color
  and thermal image fusion for object detection in automated driving,''
  \emph{Irish Machine Vision and Image Processing}, 2020.

\bibitem{mohapatra2021bevdetnet}
S.~Mohapatra, S.~Yogamani, H.~Gotzig, S.~Milz, and P.~Mader, ``Bevdetnet:
  bird's eye view lidar point cloud based real-time 3d object detection for
  autonomous driving,'' in \emph{2021 IEEE International Intelligent
  Transportation Systems Conference (ITSC)}.\hskip 1em plus 0.5em minus
  0.4em\relax IEEE, 2021, pp. 2809--2815.

\bibitem{dasgupta2022spatio}
K.~Dasgupta, A.~Das, S.~Das, U.~Bhattacharya, and S.~Yogamani,
  ``Spatio-contextual deep network-based multimodal pedestrian detection for
  autonomous driving,'' \emph{IEEE Transactions on Intelligent Transportation
  Systems}, 2022.

\bibitem{eising2021near}
C.~Eising, J.~Horgan, and S.~Yogamani, ``Near-field perception for low-speed
  vehicle automation using surround-view fisheye cameras,'' \emph{IEEE
  Transactions on Intelligent Transportation Systems}, 2021.

\bibitem{kumar2022surround}
V.~R. Kumar, C.~Eising, C.~Witt, and S.~Yogamani, ``Surround-view fisheye
  camera perception for automated driving: Overview, survey and challenges,''
  \emph{arXiv preprint arXiv:2205.13281}, 2022.

\bibitem{maddern20171}
W.~Maddern, G.~Pascoe, C.~Linegar, and P.~Newman, ``1 year, 1000 km: The oxford
  robotcar dataset,'' \emph{The International Journal of Robotics Research},
  vol.~36, no.~1, pp. 3--15, 2017.

\bibitem{kitti360_2022}
Y.~Liao, J.~Xie, and A.~Geiger, ``{KITTI}-360: A novel dataset and benchmarks
  for urban scene understanding in 2d and 3d,'' \emph{Pattern Analysis and
  Machine Intelligence (PAMI)}, 2022.

\bibitem{yogamani2019_woodscape}
S.~Yogamani, C.~Hughes, J.~Horgan, G.~Sistu, P.~Varley, D.~O'Dea,
  M.~Uric{\'a}r, S.~Milz, M.~Simon, K.~Amende, \emph{et~al.}, ``{Woodscape: A
  multi-task, multi-camera fisheye dataset for autonomous driving},'' in
  \emph{Proceedings of the IEEE/CVF International Conference on Computer Vision
  (CVPR)}, 2019, pp. 9308--9318.

\bibitem{uricar2019challenges}
M.~Uric{\'a}r, D.~Hurych, P.~Krizek, \emph{et~al.}, ``{Challenges in designing
  datasets and validation for autonomous driving},'' in \emph{Proceedings of
  the International Conference on Computer Vision Theory and Applications},
  2019.

\bibitem{dahal2021online}
A.~Dahal, V.~R. Kumar, S.~Yogamani, \emph{et~al.}, ``{An online learning system
  for wireless charging alignment using surround-view fisheye cameras},''
  \emph{IEEE Robotics and Automation Letters}, 2021.

\bibitem{rashedfisheyeyolo}
H.~Rashed, E.~Mohamed, G.~Sistu, \emph{et~al.}, ``{FisheyeYOLO: Object
  Detection on Fisheye Cameras for Autonomous Driving},'' \emph{Machine
  Learning for Autonomous Driving NeurIPSW}, 2020.

\bibitem{rashed2021generalized}
H.~Rashed, E.~Mohamed, G.~Sistu, V.~R. Kumar, C.~Eising, A.~El-Sallab, and
  S.~Yogamani, ``Generalized object detection on fisheye cameras for autonomous
  driving: Dataset, representations and baseline,'' in \emph{Proceedings of the
  IEEE/CVF Winter Conference on Applications of Computer Vision}, 2021, pp.
  2272--2280.

\bibitem{yahiaoui2019optimization}
L.~Yahiaoui, C.~Hughes, J.~Horgan, \emph{et~al.}, ``{Optimization of ISP
  parameters for object detection algorithms},'' \emph{Electronic Imaging},
  vol. 2019, no.~15, pp. 44--1, 2019.

\bibitem{dahal2019deeptrailerassist}
A.~Dahal, J.~Hossen, C.~Sumanth, G.~Sistu, K.~Malhan, M.~Amasha, and
  S.~Yogamani, ``Deeptrailerassist: Deep learning based trailer detection,
  tracking and articulation angle estimation on automotive rear-view camera,''
  in \emph{Proceedings of the IEEE/CVF International Conference on Computer
  Vision Workshops}, 2019, pp. 0--0.

\bibitem{uricar2021let}
M.~Uricar, G.~Sistu, H.~Rashed, A.~Vobecky, V.~R. Kumar, P.~Krizek, F.~Burger,
  and S.~Yogamani, ``Let's get dirty: Gan based data augmentation for camera
  lens soiling detection in autonomous driving,'' in \emph{Proceedings of the
  IEEE/CVF Winter Conference on Applications of Computer Vision}, 2021, pp.
  766--775.

\bibitem{das2020tiledsoilingnet}
A.~Das, P.~K{\v{r}}{\'\i}{\v{z}}ek, G.~Sistu, \emph{et~al.},
  ``{Tiledsoilingnet: Tile-level soiling detection on automotive surround-view
  cameras using coverage metric},'' in \emph{Proceedings of the International
  Conference on Intelligent Transportation Systems}, 2020, pp. 1--6.

\bibitem{uricar2019desoiling}
M.~Uric{\'a}r, J.~Ulicny, G.~Sistu, \emph{et~al.}, ``{Desoiling dataset:
  Restoring soiled areas on automotive fisheye cameras},'' in \emph{Proceedings
  of the International Conference on Computer Vision Workshops}.\hskip 1em plus
  0.5em minus 0.4em\relax IEEE, 2019, pp. 4273--4279.

\bibitem{cheke2022fisheyepixpro}
R.~Cheke, G.~Sistu, C.~Eising, P.~van~de Ven, V.~R. Kumar, and S.~Yogamani,
  ``Fisheyepixpro: self-supervised pretraining using fisheye images for
  semantic segmentation,'' in \emph{Electronic Imaging, Autonomous Vehicles and
  Machines Conference 2022}, 2022.

\bibitem{sobh2021adversarial}
I.~Sobh, A.~Hamed, V.~Ravi~Kumar, \emph{et~al.}, ``{Adversarial attacks on
  multi-task visual perception for autonomous driving},'' \emph{Journal of
  Imaging Science and Technology}, vol.~65, no.~6, pp. 60\,408--1, 2021.

\bibitem{dahal2021roadedgenet}
A.~Dahal, E.~Golab, R.~Garlapati, \emph{et~al.}, ``{RoadEdgeNet: Road Edge
  Detection System Using Surround View Camera Images},'' in \emph{Electronic
  Imaging}.\hskip 1em plus 0.5em minus 0.4em\relax Society for Imaging Science
  and Technology, 2021.

\bibitem{klingner2022detecting}
M.~Klingner, V.~R. Kumar, S.~Yogamani, A.~B{\"a}r, and T.~Fingscheidt,
  ``Detecting adversarial perturbations in multi-task perception,'' \emph{arXiv
  preprint arXiv:2203.01177}, 2022.

\bibitem{rashed2019motion}
H.~Rashed, A.~El~Sallab, S.~Yogamani, \emph{et~al.}, ``{Motion and depth
  augmented semantic segmentation for autonomous navigation},'' in
  \emph{Proceedings of the Computer Vision and Pattern Recognition Conference
  Workshops}, 2019, pp. 364--370.

\bibitem{dhananjaya2021weather}
M.~M. Dhananjaya, V.~R. Kumar, and S.~Yogamani, ``{Weather and Light Level
  Classification for Autonomous Driving: Dataset, Baseline and Active
  Learning},'' in \emph{Proceedings of the International Conference on
  Intelligent Transportation Systems}.\hskip 1em plus 0.5em minus 0.4em\relax
  IEEE, 2021.

\bibitem{kumar2018monocular}
V.~Ravi~Kumar, S.~Milz, C.~Witt, \emph{et~al.}, ``{Monocular fisheye camera
  depth estimation using sparse lidar supervision},'' in \emph{Proceedings of
  the International Conference on Intelligent Transportation Systems}, 2018,
  pp. 2853--2858.

\bibitem{kumar2018near}
V.~Ravi~Kumar, S.~Milz, C.~Witt, \emph{et~al.}, ``{Near-field depth estimation
  using monocular fisheye camera: A semi-supervised learning approach using
  sparse LiDAR data},'' in \emph{Proceedings of the Computer Vision and Pattern
  Recognition Conference Workshops}, vol.~7, 2018.

\bibitem{kumar2021svdistnet}
V.~R. Kumar, M.~Klingner, S.~Yogamani, \emph{et~al.}, ``{SVDistNet:
  Self-Supervised Near-Field Distance Estimation on Surround View Fisheye
  Cameras},'' \emph{Transactions on Intelligent Transportation Systems}, 2021.

\bibitem{kumar2020unrectdepthnet}
V.~Ravi~Kumar, S.~Yogamani, M.~Bach, \emph{et~al.}, ``{UnRectDepthNet:
  Self-Supervised Monocular Depth Estimation using a Generic Framework for
  Handling Common Camera Distortion Models},'' in \emph{Proceedings of the
  International Conference on Intelligent Robots and Systems}, 2020, pp.
  8177--8183.

\bibitem{kumar2020fisheyedistancenet}
V.~Ravi~Kumar, S.~A. Hiremath, M.~Bach, \emph{et~al.}, ``{Fisheyedistancenet:
  Self-supervised scale-aware distance estimation using monocular fisheye
  camera for autonomous driving},'' in \emph{Proceedings of the International
  Conference on Robotics and Automation}, 2020, pp. 574--581.

\bibitem{kumar2021fisheyedistancenet++}
V.~Ravi~Kumar, S.~Yogamani, S.~Milz, \emph{et~al.}, ``{FisheyeDistanceNet++:
  Self-Supervised Fisheye Distance Estimation with Self-Attention, Robust Loss
  Function and Camera View Generalization},'' in \emph{Electronic
  Imaging}.\hskip 1em plus 0.5em minus 0.4em\relax Society for Imaging Science
  and Technology, 2021.

\bibitem{kumar2020syndistnet}
V.~Ravi~Kumar, M.~Klingner, S.~Yogamani, \emph{et~al.}, ``{Syndistnet:
  Self-supervised monocular fisheye camera distance estimation synergized with
  semantic segmentation for autonomous driving},'' in \emph{Proceedings of the
  Workshop on Applications of Computer Vision}, 2021, pp. 61--71.

\bibitem{siam2018modnet}
M.~Siam, H.~Mahgoub, M.~Zahran, S.~Yogamani, M.~Jagersand, and A.~El-Sallab,
  ``Modnet: Motion and appearance based moving object detection network for
  autonomous driving,'' in \emph{2018 21st International Conference on
  Intelligent Transportation Systems (ITSC)}.\hskip 1em plus 0.5em minus
  0.4em\relax IEEE, 2018, pp. 2859--2864.

\bibitem{yahiaoui2019fisheyemodnet}
M.~Yahiaoui, H.~Rashed, L.~Mariotti, \emph{et~al.}, ``{FisheyeMODNet: Moving
  Object Detection on Surround-view Cameras for Autonomous Driving},'' in
  \emph{Proceedings of the Irish Machine Vision and Image Processing}, 2019.

\bibitem{mohamed2021monocular}
E.~Mohamed, M.~Ewaisha, M.~Siam, \emph{et~al.}, ``{Monocular instance motion
  segmentation for autonomous driving: Kitti instancemotseg dataset and
  multi-task baseline},'' in \emph{Proceedings of the Intelligent Vehicles
  Symposium}.\hskip 1em plus 0.5em minus 0.4em\relax IEEE, 2021, pp. 114--121.

\bibitem{tripathi2020trained}
N.~Tripathi and S.~Yogamani, ``{Trained trajectory based automated parking
  system using Visual SLAM},'' in \emph{Proceedings of the Computer Vision and
  Pattern Recognition Conference Workshops}, 2021.

\bibitem{gallagher2021hybrid}
L.~Gallagher, V.~R. Kumar, S.~Yogamani, and J.~B. McDonald, ``A hybrid
  sparse-dense monocular slam system for autonomous driving,'' in \emph{2021
  European Conference on Mobile Robots (ECMR)}.\hskip 1em plus 0.5em minus
  0.4em\relax IEEE, 2021, pp. 1--8.

\bibitem{leang2020dynamic}
I.~Leang, G.~Sistu, F.~B{\"u}rger, \emph{et~al.}, ``{Dynamic task weighting
  methods for multi-task networks in autonomous driving systems},'' in
  \emph{Proceedings of the International Conference on Intelligent
  Transportation Systems}.\hskip 1em plus 0.5em minus 0.4em\relax IEEE, 2020,
  pp. 1--8.

\bibitem{kumar2021omnidet}
V.~R. Kumar, S.~Yogamani, H.~Rashed, G.~Sitsu, C.~Witt, I.~Leang, S.~Milz, and
  P.~M{\"a}der, ``Omnidet: Surround view cameras based multi-task visual
  perception network for autonomous driving,'' \emph{IEEE Robotics and
  Automation Letters}, vol.~6, no.~2, pp. 2830--2837, 2021.

\bibitem{wrenninge2018synscapes}
M.~Wrenninge and J.~Unger, ``Synscapes: A photorealistic synthetic dataset for
  street scene parsing,'' 2018.

\bibitem{cordts2016cityscapes}
M.~Cordts, M.~Omran, S.~Ramos, T.~Rehfeld, M.~Enzweiler, R.~Benenson,
  U.~Franke, S.~Roth, and B.~Schiele, ``The cityscapes dataset for semantic
  urban scene understanding,'' 2016.

\bibitem{sekkat2022synwoodscape}
A.~R. Sekkat, Y.~Dupuis, V.~R. Kumar, H.~Rashed, S.~Yogamani, P.~Vasseur, and
  P.~Honeine, ``Synwoodscape: Synthetic surround-view fisheye camera dataset
  for autonomous driving,'' \emph{arXiv preprint arXiv:2203.05056}, 2022.

\bibitem{domain_adapt_1_2017}
\BIBentryALTinterwordspacing
G.~Csurka, \emph{Domain Adaptation in Computer Vision Applications}.\hskip 1em
  plus 0.5em minus 0.4em\relax Springer International Publishing, 2017.
  [Online]. Available: \url{http://dx.doi.org/10.1007/978-3-319-58347-1}
\BIBentrySTDinterwordspacing

\bibitem{domain_adapt_2_review_10.1007}
A.~Farahani, S.~Voghoei, K.~Rasheed, and H.~R. Arabnia, ``A brief review of
  domain adaptation,'' in \emph{Advances in Data Science and Information
  Engineering}, R.~Stahlbock, G.~M. Weiss, M.~Abou-Nasr, C.-Y. Yang, H.~R.
  Arabnia, and L.~Deligiannidis, Eds.\hskip 1em plus 0.5em minus 0.4em\relax
  Cham: Springer International Publishing, 2021, pp. 877--894.

\bibitem{domain_adapt_3_review_technologies8020035}
\BIBentryALTinterwordspacing
M.~Toldo, A.~Maracani, U.~Michieli, and P.~Zanuttigh, ``Unsupervised domain
  adaptation in semantic segmentation: A review,'' \emph{Technologies}, vol.~8,
  no.~2, 2020. [Online]. Available: \url{https://www.mdpi.com/2227-7080/8/2/35}
\BIBentrySTDinterwordspacing

\bibitem{reinhard2001}
E.~Reinhard, M.~Ashikhmin, B.~Gooch, and P.~Shirley, ``Color transfer between
  images,'' \emph{IEEE Computer Graphics and Applications}, vol.~21, pp.
  34--41, 10 2001.

\bibitem{lyu2020learning}
Q.~Lyu, M.~Chen, and X.~Chen, ``Learning color space adaptation from synthetic
  to real images of cirrus clouds,'' 2020.

\bibitem{Imbusch_2022}
\BIBentryALTinterwordspacing
B.~T. Imbusch, M.~Schwarz, and S.~Behnke, ``Synthetic-to-real domain adaptation
  using contrastive unpaired translation,'' in \emph{2022 {IEEE} 18th
  International Conference on Automation Science and Engineering
  ({CASE})}.\hskip 1em plus 0.5em minus 0.4em\relax {IEEE}, aug 2022. [Online].
  Available: \url{https://doi.org/10.1109%2Fcase49997.2022.9926640}
\BIBentrySTDinterwordspacing

\bibitem{sankaranarayanan2018learning}
S.~Sankaranarayanan, Y.~Balaji, A.~Jain, S.~N. Lim, and R.~Chellappa,
  ``Learning from synthetic data: Addressing domain shift for semantic
  segmentation,'' 2018.

\bibitem{hoffman2017cycada}
J.~Hoffman, E.~Tzeng, T.~Park, J.-Y. Zhu, P.~Isola, K.~Saenko, A.~A. Efros, and
  T.~Darrell, ``Cycada: Cycle-consistent adversarial domain adaptation,'' 2017.

\bibitem{ganin2015unsupervised}
Y.~Ganin and V.~Lempitsky, ``Unsupervised domain adaptation by
  backpropagation,'' 2015.

\bibitem{vu2019advent}
T.-H. Vu, H.~Jain, M.~Bucher, M.~Cord, and P.~Pérez, ``Advent: Adversarial
  entropy minimization for domain adaptation in semantic segmentation,'' 2019.

\bibitem{wang2020classes}
H.~Wang, T.~Shen, W.~Zhang, L.~Duan, and T.~Mei, ``Classes matter: A
  fine-grained adversarial approach to cross-domain semantic segmentation,''
  2020.

\bibitem{ramachandran2021woodscape}
S.~Ramachandran, G.~Sistu, J.~McDonald, and S.~Yogamani, ``Woodscape fisheye
  semantic segmentation for autonomous driving--cvpr 2021 omnicv workshop
  challenge,'' \emph{arXiv preprint arXiv:2107.08246}, 2021.

\bibitem{ramachandran2022woodscape}
S.~Ramachandran, G.~Sistu, V.~R. Kumar, J.~McDonald, and S.~Yogamani,
  ``Woodscape fisheye object detection for autonomous driving -- cvpr 2022
  omnicv workshop challenge,'' 2022.

\bibitem{he2015deep_resnet}
K.~He, X.~Zhang, S.~Ren, and J.~Sun, ``Deep residual learning for image
  recognition,'' in \emph{2016 IEEE Conference on Computer Vision and Pattern
  Recognition (CVPR)}, 2016, pp. 770--778.

\bibitem{pdtraining2023}
\BIBentryALTinterwordspacing
M.~Galarnyk, N.~Cibik, O.~Maher, and P.~Thomas, ``{Synthetic Data Best
  Practices for Perception Applications},'' Parallel Domain Blog, 2023.
  [Online]. Available: \url{Available at
  https://paralleldomain.com/synthetic-data-best-practices-for-perception-applications}
\BIBentrySTDinterwordspacing

\bibitem{cvprsdas2023}
P.~Thomas, ``On the importance of label domain gaps,'' Talk presented at the
  CVPR 2023 Workshop: Synthesized Data and Augmentation Strategies (SDAS),
  2023, available at: \url{https://sites.google.com/view/sdas2023/}.

\bibitem{cheng2021mask2former}
B.~Cheng, I.~Misra, A.~G. Schwing, A.~Kirillov, and R.~Girdhar,
  ``Masked-attention mask transformer for universal image segmentation,'' in
  \emph{2022 IEEE/CVF Conference on Computer Vision and Pattern Recognition
  (CVPR)}, 2022, pp. 1280--1289.

\bibitem{liu2021swin}
Z.~Liu, Y.~Lin, Y.~Cao, H.~Hu, Y.~Wei, Z.~Zhang, S.~Lin, and B.~Guo, ``Swin
  transformer: Hierarchical vision transformer using shifted windows,'' 2021.

\bibitem{liu_convnext}
Z.~Liu, H.~Mao, C.-Y. Wu, C.~Feichtenhofer, T.~Darrell, and S.~Xie, ``A convnet
  for the 2020s,'' in \emph{2022 IEEE/CVF Conference on Computer Vision and
  Pattern Recognition (CVPR)}, 2022, pp. 11\,966--11\,976.

\bibitem{internimage_Wang_2023_CVPR}
W.~Wang, J.~Dai, Z.~Chen, Z.~Huang, Z.~Li, X.~Zhu, X.~Hu, T.~Lu, L.~Lu, H.~Li,
  X.~Wang, and Y.~Qiao, ``Internimage: Exploring large-scale vision foundation
  models with deformable convolutions,'' in \emph{Proceedings of the IEEE/CVF
  Conference on Computer Vision and Pattern Recognition (CVPR)}, June 2023, pp.
  14\,408--14\,419.

\bibitem{segmenter_Strudel_2021_ICCV}
R.~Strudel, R.~Garcia, I.~Laptev, and C.~Schmid, ``Segmenter: Transformer for
  semantic segmentation,'' in \emph{Proceedings of the IEEE/CVF International
  Conference on Computer Vision (ICCV)}, October 2021, pp. 7262--7272.

\bibitem{adamw_KingmaB14}
D.~P. Kingma and J.~Ba, ``Adam: {A} method for stochastic optimization,'' in
  \emph{International Conference on Learning Representations (ICLR)}, 2015.

\bibitem{polylr_Zhang2020kDecayJA}
\BIBentryALTinterwordspacing
T.~Zhang and W.~Li, ``kdecay: Just adding k-decay items on learning-rate
  schedule to improve neural networks,'' 2020. [Online]. Available:
  \url{https://api.semanticscholar.org/CorpusID:247597102}
\BIBentrySTDinterwordspacing

\bibitem{upper_xiao2018unified}
T.~Xiao, Y.~Liu, B.~Zhou, Y.~Jiang, and J.~Sun, ``Unified perceptual parsing
  for scene understanding,'' in \emph{European Conference on Computer
  Vision}.\hskip 1em plus 0.5em minus 0.4em\relax Springer, 2018.

\bibitem{zhang2021knet}
W.~Zhang, J.~Pang, K.~Chen, and C.~C. Loy, ``{K-Net: Towards} unified image
  segmentation,'' in \emph{NeurIPS}, 2021.

\bibitem{fda_9157228}
Y.~Yang and S.~Soatto, ``Fda: Fourier domain adaptation for semantic
  segmentation,'' in \emph{2020 IEEE/CVF Conference on Computer Vision and
  Pattern Recognition (CVPR)}, 2020, pp. 4084--4094.

\end{thebibliography}
\end{document}